\documentstyle[aclap,psfig]{article}
\author{Simone Teufel\\IMS-CL Institut f\"ur Maschinelle
  Sprachverarbeitung -- Computerlinguistik\\Universit\"at
  Stuttgart\\\fbox{Workshop SIGDAT (EACL95)}}
\title{A Support Tool for Tagset Mapping}
\begin{document}
\maketitle
\begin{abstract}Many different tagsets are used in existing corpora;
  these tagsets vary according to the objectives of specific projects
  (which may be as far apart as robust parsing vs.  spelling
  correction). In many situations, however, one would like to have
  uniform access to the linguistic information encoded in corpus
  annotations without having to know the classification schemes in
  detail. This paper describes a tool which maps unstructured
  morphosyntactic tags to a constraint-based, typed, configurable
  specification language, a ``standard tagset''.  The mapping relies
  on a manually written set of mapping rules, which is automatically
  checked for consistency.  In certain cases, unsharp mappings are
  unavoidable, and noise, i.e.  groups of word forms {\sl not} conforming
  to the specification, will appear in the output of the mapping.  The
  system automatically detects such noise and informs the user about it.

  The tool has been tested with rules for the UPenn tagset \cite{up}
  and the SUSANNE tagset \cite{garside}, in the framework of the
  EAGLES\footnote{LRE project EAGLES, cf.  \cite{eagles}.} validation
  phase for standardised tagsets for European languages.
\end{abstract}

\section{Motivation}
Tagsets used in existing corpora have usually been designed to
satisfy the needs of specific projects. A tagset used for robust
parsing will tend to stress distributional properties, whereas a
corpus within a lexical resource specially designed for human
interaction (which might include a human oriented dictionary) will
most likely distinguish word classes along traditional linguistic
lines.

The tool described in this paper performs tagset mapping with manually
written rules to introduce a standardised morphosyntactic tagset.
Standardisation of tagsets has been a goal of some contemporary
projects (e.g.  \cite{eagles} and the Text Encoding Initiative
\cite{tei}); at the same time, it has been the object of much
controversy because of the obvious advantages of tailoring tagsets to
project needs. Looking at the problem from a larger perspective than
that of isolated projects, a uniform tagset has the following
advantages:
\begin{itemize}
\item {\bf Objectivisation and standardisation of similar
    information}: Millions of words have been analysed in the past,
  using different annotation schemes. Especially the manually analysed
  linguistic data is expensive to produce and extremely valuable. With
  a standardised tagset, linguistic information from different corpora
  of the same language can be {\sl reused} and thus merged into a
  large data base.  Such data bases improve the performance of
  statistical methods and are a useful resource for the production of
  balanced corpora.

\item {\bf Shared use of language resources}: Corpus manipulation
  tools such as retrieval tools can be applied to merged resources in
  a uniform format without much customisation.  As well, users of
  these tools will find it easier to work with a corpus tagged in a
  standardised tagset. Now,  they have to memorize only {\em one} scheme
  of tag classes (class names, class semantics, exceptions), as opposed
  to several schemes for several corpora before.

\item {\bf Comparison of annotation schemes}: A comparison of the
  granularity and degree of similarity of tagsets can be carried out
  more objectively, once the mapping results are available. The
  validation of the suggestions of the LRE-project EAGLES is an
  application in this field.
\end{itemize}

We believe that standards are important for the linguistic community,
especially from the point of view of reusablility.

\begin{figure*}[ht]
  \centerline{\psfig{file=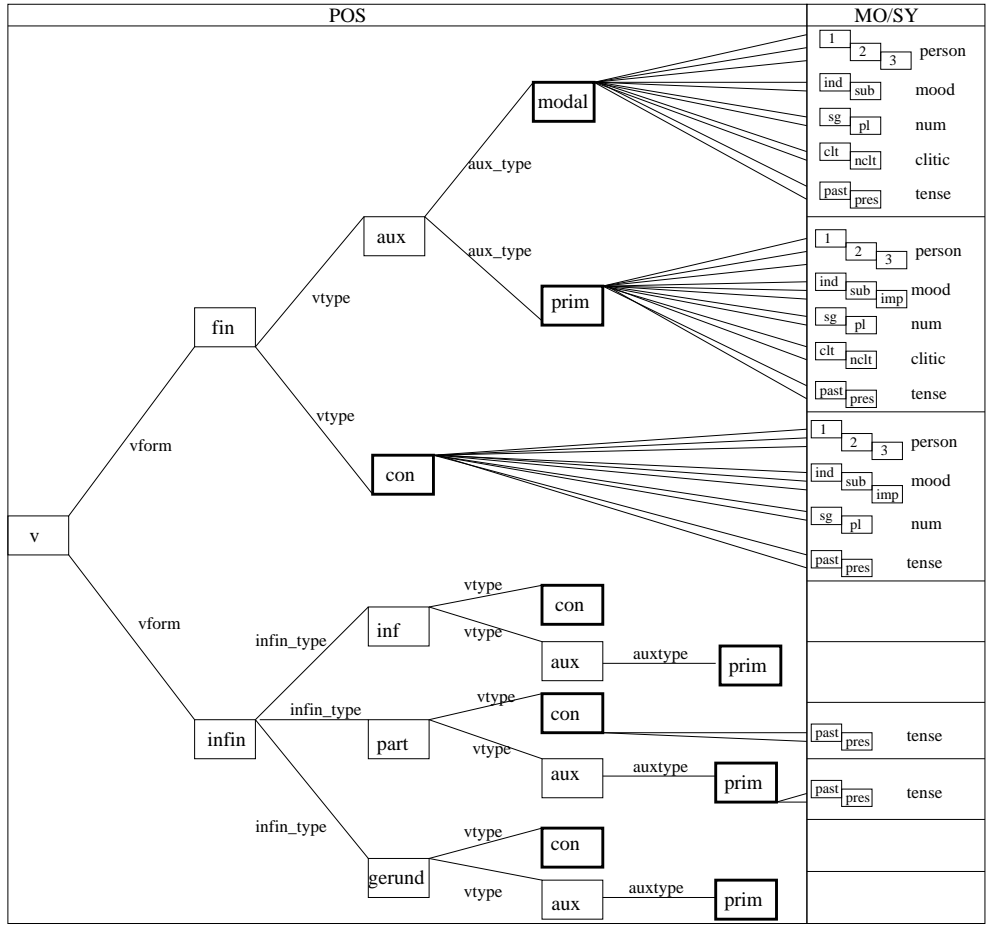,width=13cm}}
\caption{Detail of the type graph (verbs)}
\label{tgraph}
\end{figure*}

Of course, there are limits: proposals for standard tagsets should be
regarded as approaches towards a neutral platform between projects and
different theories, rather than as ready-made tagsets that will never
be changed. It is important indeed that standards and their support
tools be flexible about possible extensions and improvements.

The more general problem of retagging has been approached with tools
like ICA \cite{ica}, a public domain retagging tool which uses SGML as
interlingua\footnote{Many other retagging tools are available in
  the SGML world.}. We also know of current work at Leeds University
on mapping tagsets, though this work is concerned with the mapping of
syntactic structure encoded in corpora \cite{atwell}.

\section{A standardised tagset}
When designing the architecture of a standardised tagset, we
implemented the following constructs as they provide
considerable advantages compared to the the traditional flat word
labels.
\begin{itemize}
\item As the tagset is {\bf constraint-based}, a flexible
  generalisation is possible over all atomic constraints and
  combinations of constraints\footnote{ Constraints are expressed as
    attribute-value-pairs.}. As a formal grammar\footnote{We used a
    grammar for Boolean expressions with the usual precedences.} is
  used to define syntactically well-formed specifications of word
  forms, we can regard our standard tagset as a {\sl specification
    language}.

  Example: The specification {\tt [pos = v \& vtype = aux \& pers = 3]}
  denotes 3rd person auxiliary verbs.

\item The tagset is also {\bf typed}, which adds to the naturalness of
  the specifications of wordforms and helps discover semantic errors
  in specifications (inconsistent combinations of features, wrong
  values for features).  In our implementation, we follow the
  closed-world-assumption, which leads to a coherent interpretation
  for underspecified and/or negated descriptions.

  Example: {\tt [pos = v \& (vform = fin $|$ \\ case != gen)]} is a
  syntactically correct,
\begin{figure*}[ht]
  \centerline{\psfig{file=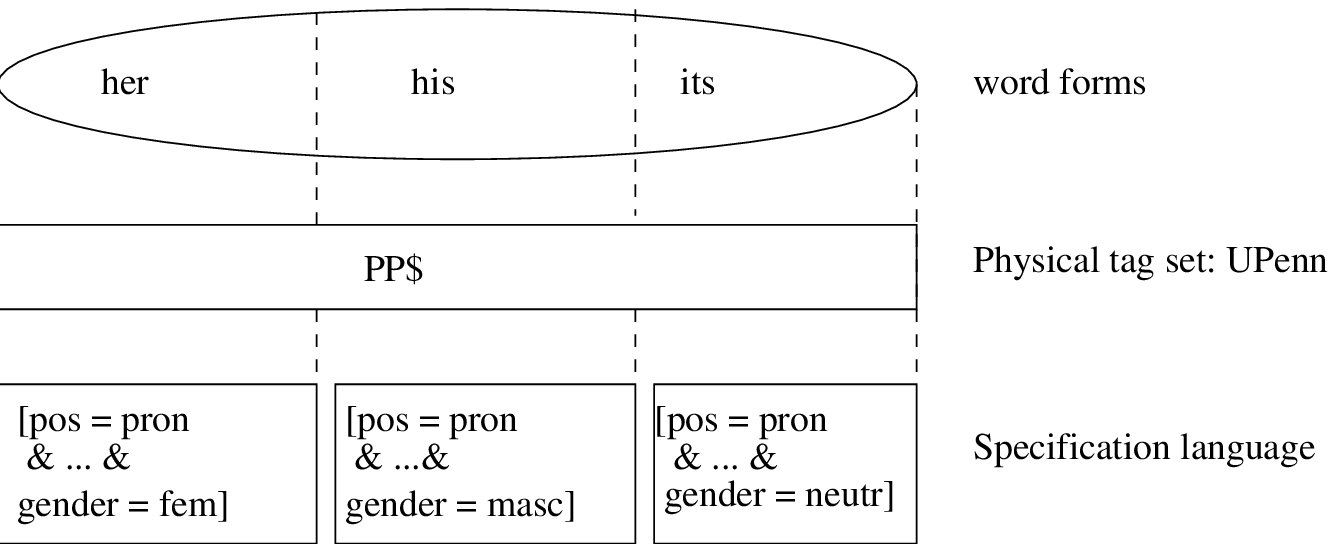,width=9cm}}
\caption{\fbox{1:n} 1 class of physical tagset $\leftrightarrow$ n classes of
specification language}
\label{1:n}
\end{figure*}

\begin{figure*}[ht]
  \centerline{\psfig{file=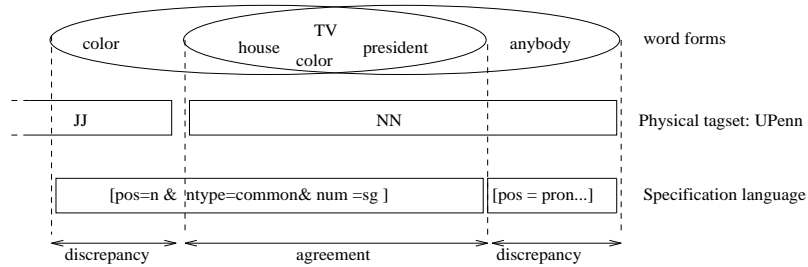,width=11cm}}
\caption{\fbox{n:m} n classes of physical tagset $\leftrightarrow$ m classes of
specification language}
\label{n:m}
\end{figure*}

 but
  ill-typed specification, as the Types {\tt v} (Verb) and {\tt gen}
  (Genitive) are not type compatible.

\item The tagset can be easily {\bf modified} because its manually
  written definition is compiled into a system internal
  format.\footnote{The system is implemented in {\sc Prolog}, and the
    definition can be spelled out as a structured {\sc Prolog} fact.}
  As the design of a tagset involves a cycle with feedback phases,
  including manual tagging and the writing of guidelines\footnote{The
    guidelines document is a very important resource for manual
    taggers as well as for users of the corpus data, as it provides
    the semantics of the tag classes.}, there will be frequent
  modifications to the tagset, especially in the initial phase.
\end{itemize}

The EAGLES expert group (cf. \cite{moca}) suggested an inventory of
features and values for a standardised morphosyntactic tagset for
European\footnote{English, French, Greek, German, Dutch, Portugese,
  Spanish, Italian, Danish.} languages; there are different layers,
depending on language specificity as well as on application
specificity.  For the design of a standardised tagset in a specific
language, relevant features and values are to be chosen from the
inventory. Fig.  \ref{tgraph} shows a detail of the tentative English
tagset we designed and used for our tests.  The type relations are
divided into hierarchical (POS) features and non-hierarchical features
(MO/SY).
%Hierarchical features devide the object space only according
%to one dimension, whereas non-hierarchical features devide the object
%space according to several dimensions if combined: in these cases, the
%Cartesian product of the values designates the most specific types.

\section{Tag mapping: the problems}
Mapping tags of an existing, flat-labeled tagset\footnote{We call
  such a tagset {\sl physical tagset} because its tags are
  actually annotated in an existing corpus, in contrast to the derived
  tags of the specification language.} or source annotation scheme to
tags of a specification language (target annotation scheme) is an
instance of the retagging problem. It is straightforward only in the
trivial cases 1:1 (renaming) and n:1. In the latter case, the physical
tagset makes finer distinctions than the target annotation scheme.
This case introduces no problem for the mapping itself even if not all
information contained in the corpus can be accessed.  Unfortunately, what we
usually find in the mapping business is a mixture of two more
problematic cases:

\fbox{1:n} The physical tagset cannot support a distinction intended
by the specification language, e.g. as the distinction {\tt gender} in
fig. \ref{1:n}. Therefore, there is a lack of information: the corpus
annotation does not provide the wanted distinction.

\fbox{n:m} There is an overlap between tag classes, as illustrated in
fig. \ref{n:m}. In the example case, the source annotation scheme
includes special indefinite pronouns like {\it anybody} into the
normal common nouns, whereas some word forms ({\it color}) are
(wrongly!) tagged as adjectives in the source annotation scheme but as
common nouns in the target annotation scheme.

\section{Mapping Rules}
We opted for symbolic mapping rules\footnote{ We also thought about
  having a program deduce mapping rules from a corpus. The automatic
  learning of tag correspondences, at least on a semiautomatic basis,
  seems possible with standard statistical means (e.g. HMM based
  learning algorithms).

  However, the amount of data needed for such an enterprise (a large training
  corpus, (manually) annotated in both source and target annotation
  scheme) made us vote for the symbolic approach.} and designed two
kinds of mapping rules to deal with the discrepancies indicated above.
\begin{itemize}
\item {\bf Class coverage rules} describe a correspondence of source
  and target annotation classes\footnote{These rules are used in cases 1:1,
n:1, 1:n
  and in the agreement area of case n:m.}. The rule format is as follows:
  for each physical tag, the equivalent expression in the specification
  language is named.

Example:  {\tt \small [pos = 'NN']    =$>$ \\ \hspace*{5mm}\mbox{[n \& ( common
\& sg  $|$
mass ) ].}}

The word forms that are annotated with the physical tag {\tt NN} are ``common
singular nouns or mass nouns'' in the terms of the specification
language.

  \begin{figure*}[ht]
  \centerline{\psfig{file=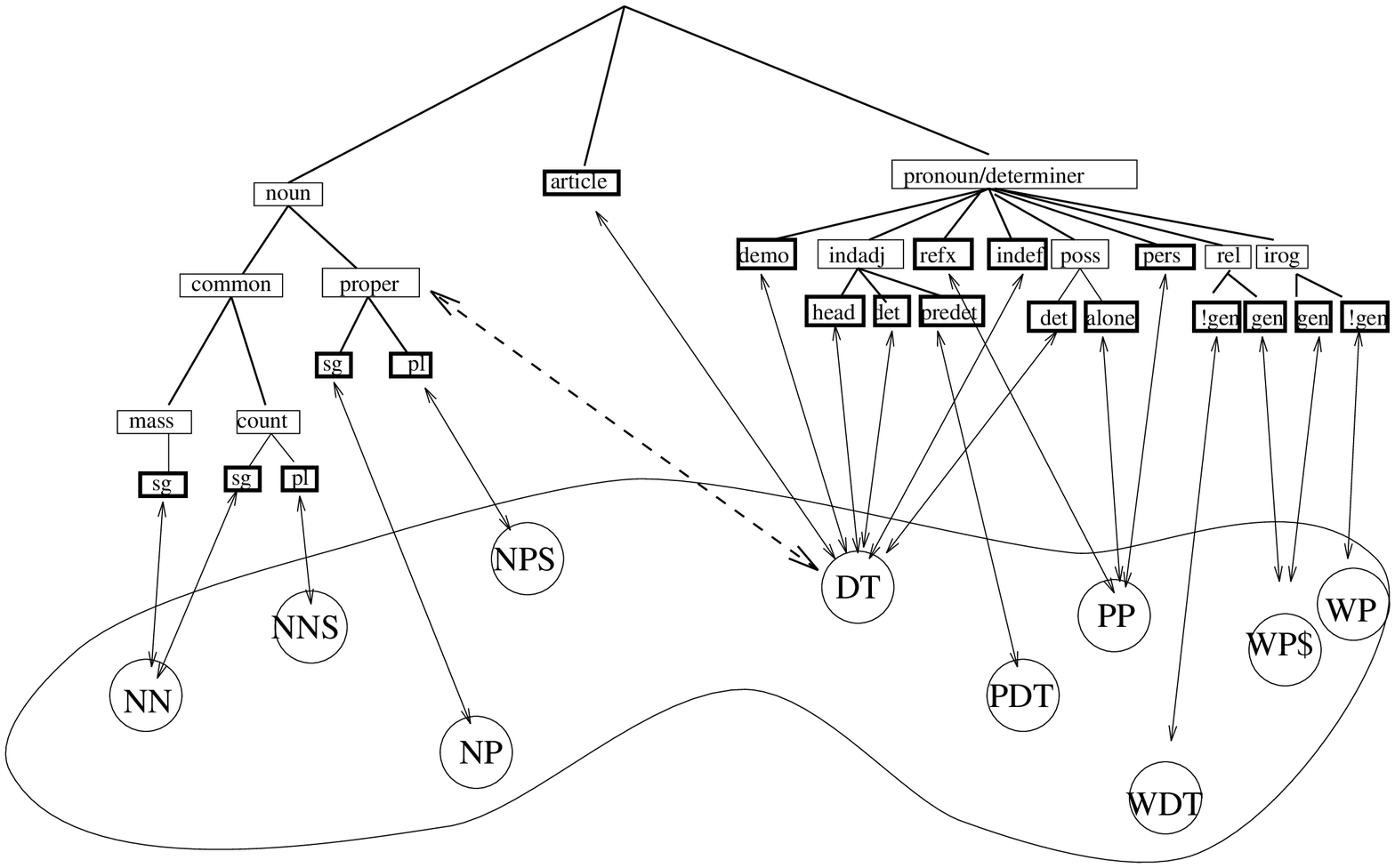,width=\textwidth}}
  \caption{Detail of the MTree for the UPenn annotation scheme}
  \label{mtree}
  \end{figure*}

\item The {\bf exception lexicon} provides a treatment of the
  individual discrepancy areas of case n:m, in order to deal with
  noise from unsharp mappings\footnote{This solution accounts for
    lexical exceptions only. Contextual discrepancies like the
    decision to tag a certain wordform like {\it that} in one class or
    in several classes (demonstrative pronoun or conjunction or
    relative pronoun) are not dealt with in this work as this includes
    a new disambiguation run (pos-tagging)}. Specific lexical items
  can be reclassified, i.e.  their standard mapping can be overridden.
  (Notation: the sign $<<$ stands for ``out of'') They can be
  reclassified in a different target annotation scheme class instead
  (sign $>>$ stands for ``into'').

  Example: The following exception lexicon entry expresses that the
  target tag for wordforms {\sl anybody, nothing \dots} in fig.
  \ref{n:m} should not be the standard reading for {\tt NN} ( common
  singular nouns or mass nouns), but should be described as an  indefinite
  pronoun relating to persons.

{\tt \small [anybody,
    nothing, something, anything] $<<$ [pos = `NN`] $>>$ [pos=pron \&
    antec=prs \& type=indef].}
\end{itemize}
  The exception lexicon lookup takes place after the mapping of the
  class coverages. For more details, see \cite{teufel}.
\section{Mtree: Internal representation\label{cc}}
After the compilation of the mapping rules, the system keeps the
information in a data structure called an MTree (mapping tree), see
fig.  \ref{mtree}, which shows the verb mappings for UPenn. There is
an MTree for each physical tagset regarded.  MTrees contain a subset
of the information contained in the type graph (see fig. \ref{ttree}),
namely only those distinctions of the original type graph that are
distinguishable in the physical tagset. The new terminals (boxes with
thick lines in fig. \ref{mtree}) in this pruned type graph correspond
to physical tags (encircled tag names).

Within the rule set, the system keeps track of consistency. Warnings
are issued in case of one of the following inconsistencies which might
occur during the construction of an MTree:
\begin{itemize}
\item {\bf definition holes}: Either target or source annotation
  schemes are not covered by a mapping rule (classes have been
  forgotten by the person writing the mapping rules).
\item {\bf nondisjunctiveness of classes}: A target annotation class
  has several source annotation correspondences. Although this might
  be an instance of case n:1, a warning is issued, because most such cases
  occur due to a conceptual error.
\item {\bf hierarchical inconsistency}: Instead of keeping a clear
  distinction between terminal classes and nonterminal classes, an odd
  mapping assigns terminal status to ancestors of classes that are
  terminals themselves. In fig. \ref{mtree}, the correspondence
  specified by the dashed arrow introduces a hierarchical
  inconsistency, as it assigns a physical tag ({\tt VBN}) to a class
  ({\tt con}) that cannot be terminal because its daughters ({\tt
    past} and {\tt pres}) already are.
\end{itemize}

\section{System Support}

System support includes
\begin{itemize}
\item Compilation of the tagset definition: useful for tagsets with many
  non-hierarchical, i.e.  combinatory features (which would have to be
  multiplied out manually otherwise.)
\item Compilation of mapping rules: consistency checks (cf. section \ref{cc}).
\item Interpretation of specifications: Each specification is
  syntactically and semantically checked, and the corresponding (set
  of) physical tag(s) is computed, using the MTree information.  Due
  to 1:n and/or n:m cases (unsharp mapping), there can be noise (i.e.
  groups of word forms which do {\sl not} conform to the
  specification) in the output.  In these cases, the system
  anticipates the noise to be expected and informs the user. Warnings
  about noise are essential for a correct interpretation of the
  output.

  Noisy word classes can be deduced from the MTree: In the MTree given
  in fig.  \ref{mtree}, we can see that target specification {\tt inf}
  (infinitives) will always induce noise from finite forms, namely
  subjunctive and imperative forms, because the physical class {\tt
    VB} does not distinguish between these groups (case 1:n).
\end{itemize}

\section{Results and Outlook} For test purposes, we wrote
mapping rules for the UPenn and SUSANNE tagsets. The number of
coverage rules is
equivalent to the number of physical tags. Rules are easy to
formulate, once users have got used to the class semantics of the
standard tag set. Information input are tagging guidelines, if the
source annotation scheme comes with a comprehensive description of the
intended class semantics\footnote{\cite{up-gl} provides tagging
  guidelines for the UPenn corpus, \cite{garside} for the SUSANNE
  corpus.}, or corpus queries otherwise, which is more time consuming.

We wrote exemplary exception lexicon entries for auxiliary verbs and
some for noun exceptions, but more work can be put into the exception
lexicon to improve the accuracy in the lexically determinable cases of
discrepancies.

Apart from being used for the validation of the EAGLES standard for
English and German, the tool has been integrated into a corpus query system
(Christ 94, Schulze 94) to allow for ``more abstract'' and corpus
independent queries. A typical query (content verbs in infinitive or
primary auxiliaries in past tense) to a specific corpus (here: UPenn) looks
like this:

\begin{verbatim}
Query> [(vtype=con & vform=inf) |
        (vtype=prim & tense=past)].

%% warning:  Noise from [con & fin & imp]
%%             and from [con & fin & sub]
%%              (Due to tag "VB")!

[((pos = "VB" & word != "be|do|have")|
(pos = "VBD" & word = "was|were|had|did")|
(pos = "VBN" & word = "been|had|done"))]

\end{verbatim}
We get the information that the system will query for tags {\tt VB, VBD,
VBN} (with lexical constraints) in the UPenn corpus; however, we must
expect to find {\sl finite} content verbs (namely imperative and subjunctive
forms) in our output (1:n case).

It would be particularly interesting to explore ways
of how to use an MRD to build an exception lexicon automatically,
which is especially useful for closed word classes.

Another interesting case are multi-word tags and discrepancies with
respect to the assignment of word boundaries
(tokenising).\footnote{For an exhaustive survey of multi-word
  phenomena, see \cite{multi}.} Compare the following cases (UPenn
tokenising and tagging):

\begin{itemize}
\item \underline{Peter/NP \hspace{0.7mm} 's/POS}\,  house
\item \underline{he/PP \hspace{0.7mm} 's/VBZ} \, not at home
\end{itemize}

In our opinion, {\it Peter's\/} should be regarded as one nominal item
(with {\tt genitive} as value for the {\tt case} attribute),
whereas {\it he} and {\it 's\/}
should be kept as two words.  We are thinking about designing a rule
construct to express this kind of word bundelling with conditional features.

\end{document}